\documentclass[letterpaper, 10 pt, journal, twoside]{IEEEtran}

\usepackage{cite}
\usepackage{graphicx}
\usepackage{amsmath,amssymb}
\usepackage{xcolor}
\usepackage{siunitx}
\usepackage{bm}
\usepackage{hyperref}
\usepackage[font=small]{caption, subcaption}
\usepackage{microtype} 

\usepackage{mathtools}

\newcommand{\R}{\mathtt{R}}

\newcommand{\errr}{\tilde{\boldsymbol{\theta}}}
\newcommand{\errp}{\boldsymbol{\delta}\tilde{\boldsymbol{p}}}
\newcommand{\errv}{\boldsymbol{\delta}\tilde{\boldsymbol{v}}}
\newcommand{\errb}{\boldsymbol{\delta}\tilde{\boldsymbol{b}}}

\newcommand{\sof}{\mathfrak{so}}

\newcommand{\Cov}{\boldsymbol{\Sigma}}

\newcommand{\rotvel}{\boldsymbol{\omega}}
\newcommand{\acc}{\boldsymbol{a}}

\newcommand{\State}{\boldsymbol{X}}
\newcommand{\uncertvec}{\hat{\boldsymbol{u}}} 

\newcommand{\rot}{\ensuremath{\prescript{\text{w}}{\text{i}}{\boldsymbol{\R}}}}
\newcommand{\tran}{\ensuremath{\prescript{\text{w}}{}{\boldsymbol{p}}}}
\newcommand{\speed}{\ensuremath{\prescript{\text{w}}{}{\boldsymbol{v}}}}

\newcommand{\rotest}{\ensuremath{\prescript{\text{w}}{\text{i}}{\hat{\boldsymbol{\R}}}}}
\newcommand{\tranest}{\ensuremath{\prescript{\text{w}}{}{\hat{\boldsymbol{p}}}}}
\newcommand{\speedest}{\ensuremath{\prescript{\text{w}}{}{\hat{\boldsymbol{v}}}}} 
\newcommand{\dispest}{\ensuremath{\hat{\boldsymbol{d}}}}

\newcommand{\dispmeas}{\ensuremath{\boldsymbol{d}}}
\newcommand{\grav}{\prescript{\text{w}}{}{\boldsymbol{g}}}
\newcommand{\bias}{\boldsymbol{b}}

\newcommand{\Js}{\boldsymbol{A}} 
\newcommand{\Ju}{\boldsymbol{B}} 

\newcommand{\clst}{\boldsymbol{\xi}} 
\newcommand{\cust}{\boldsymbol{s}} 
\newcommand{\clstexp}{\xi}  

\newcommand{\brw}{\boldsymbol{\eta}}
\DeclarePairedDelimiter\abs{\lvert}{\rvert}%
\DeclarePairedDelimiter\norm{\lVert}{\rVert}%

\makeatletter
\let\oldabs\abs
\def\abs{\@ifstar{\oldabs}{\oldabs*}}
\let\oldnorm\norm
\def\norm{\@ifstar{\oldnorm}{\oldnorm*}}
\makeatother

\newcommand{\RONINMETHOD}{\textsc{3d-ronin}\xspace}
\newcommand{\FILTERMETHOD}{\textsc{tlio}\xspace}

\usepackage{mathtools}

\DeclarePairedDelimiter\floor{\lfloor}{\rfloor}
\input{macros/macro_alphabet.tex}

\begin{document}

© 2020 IEEE. Personal use of this material is permitted. Permission from IEEE must be
obtained for all other uses, in any current or future media, including
reprinting/republishing this material for advertising or promotional purposes, creating new
collective works, for resale or redistribution to servers or lists, or reuse of any copyrighted
component of this work in other works.
\thispagestyle{empty}
\clearpage
\setcounter{page}{1}
\newpage

\title{TLIO: Tight Learned Inertial Odometry}

\author{Wenxin Liu$^{1,2}$, David Caruso$^{2}$, Eddy Ilg$^{2}$, Jing Dong$^{2}$, Anastasios I. Mourikis$^{2}$,

Kostas Daniilidis$^{1}$, Vijay Kumar$^{1}$, and Jakob Engel$^{2}$%

\thanks{
} 
\thanks{$^{1} $GRASP Lab, University of Pennsylvania, Philadelphia, USA {\tt\footnotesize wenxinl@seas.upenn.edu}}
\thanks{$^{2} $Facebook Reality Labs, Redmond, USA {\tt\footnotesize dcaruso@fb.com}}
}

\markboth{IEEE Robotics and Automation Letters. Preprint Version. Accepted June, 2020}
{Liu \MakeLowercase{\textit{et al.}}: TLIO: Tight Learned Inertial Odometry} 

\maketitle

\begin{abstract}
In this work we propose a tightly-coupled Extended Kalman Filter framework for IMU-only state estimation. Strap-down IMU measurements provide relative state estimates based on IMU kinematic motion model. However the integration of measurements is sensitive to sensor bias and noise, causing significant drift within seconds. Recent research by Yan et al. (RoNIN) and Chen et al. (IONet) showed the capability of using trained neural networks to obtain accurate 2D displacement estimates from segments of IMU data and obtained good position estimates from concatenating them. This paper demonstrates a network that regresses 3D displacement estimates and its uncertainty, giving us the ability to tightly fuse the relative state measurement into a stochastic cloning EKF to solve for pose, velocity and sensor biases. We show that our network, trained with pedestrian data from a headset, can produce statistically consistent measurement and uncertainty to be used as the update step in the filter, and the tightly-coupled system outperforms velocity integration approaches in position estimates, and AHRS attitude filter in orientation estimates. 

{\small \bf \centering Video materials and code can be found on our project page: \href{http://cathias.github.io/TLIO/}{\color{cyan} cathias.github.io/TLIO}\par}
\end{abstract}

\begin{IEEEkeywords}
Localization, AI-Based Methods, Pedestrian Dead Reckoning, Inertial State Estimation
\end{IEEEkeywords}

\section{Introduction} \label{sec:intro}

\IEEEPARstart{V}{isual}-Inertial Navigation Systems (VINS) in recent years have seen tremendous success, enabling a wide range of applications from mobile devices to autonomous systems. Using only light-weight cameras and Inertial Measurement Units (IMUs), VINS achieve high accuracy in tracking at low cost and provide one of the best solutions for localization and navigation on constrained platforms. With the unique combination of these advantages, VINS have been the de facto standard for demanding applications such as Virtual/Augmented Reality (VR/AR) on mobile phones or headsets.

Despite the impressive performance of state-of-the-art VINS algorithms, demands from consumer AR/VR products are posing new challenges on state estimation, pushing the research frontier. Visual-Inertial Odometry (VIO) largely relies on consistent image tracking, leading to failure cases under extreme lighting or camera blockage conditions, such as in a dark room or inside a pocket. High frequency image processing makes power consumption a bottleneck for sustainable long-term operations. 
In addition, widespread camera usage carries privacy implications.
Targeting an alternative to the state-of-the-art VINS algorithms for pedestrian applications, this paper focuses on consumer grade IMU-only state estimation problem known as strap-down Inertial Navigation System (INS) or Dead Reckoning~\cite{groves2015principles}.

\begin{figure}[t!]
	\centering
	\includegraphics[width=1\linewidth]{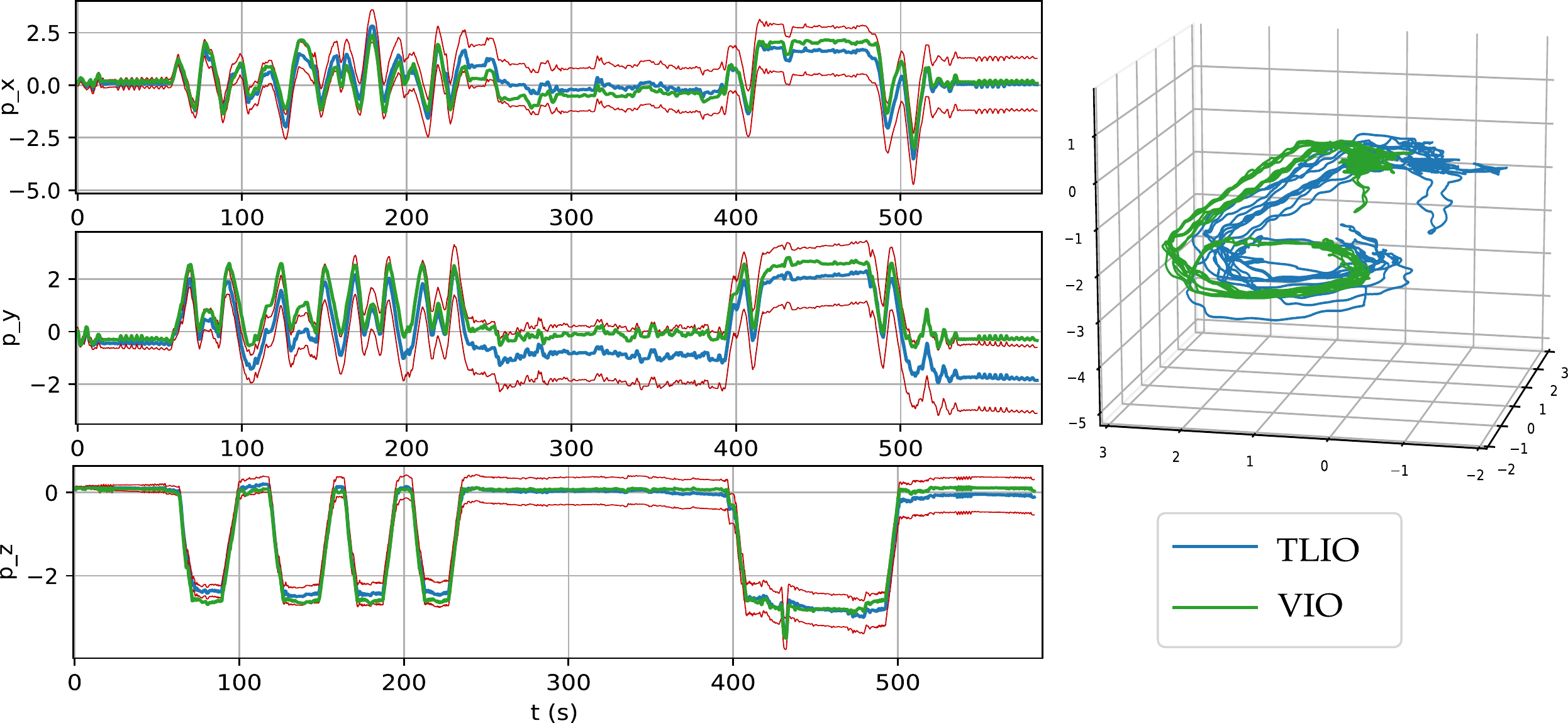}
	\caption{An example 3D trajectory estimated by \FILTERMETHOD from a single MEMS IMU, showing the user repeatedly walking up and down a staircase. Our method estimates full orientation and translation in 3D: On the left, we show the estimated x,y,z position (blue) as well as associated uncertainties ($\pm 3 \sigma$ in red). On the right, we show the resulting 3D trajectory. For both plots, we show the output of state-of-the-art visual-inertial odometry (green) for comparison. 
	}
	\label{fig:overview}
\end{figure}

Inertial Pedestrian Dead Reckoning (PDR) has gained increasing interest with the price decline and widespread usage of MEMS sensors in the past two decades~\cite{harle2013survey}. 
Strap-down pedestrian IMU navigation faces the challenge of the accumulation of sensor errors since the IMU kinematic model provides only relative state estimates. 
To compensate for the errors and reduce drift without the aid of other sensors or floor maps, the existing approaches rely on the prior knowledge of human walking motion, in particular, steps. 
One way to make use of steps is Zero-velocity UPdaTe (ZUPT)~\cite{foxlin2005pedestrian}. It detects when the foot touches the ground to generate pseudo-measurement updates in an Extended Kalman Filter (EKF) framework to calibrate IMU biases. However, it only works well when the sensor is attached to the foot where step detection is obvious. Another category is step counting~\cite{brajdic2013walk} which does not require the sensor to be attached to the foot. Such systems consist of multiple submodules: the identification and classification of steps, data segmentation and step length prediction, all of which require heavy tuning of hand-designed heuristics or machine learning.

In parallel, recent research advances (IONet\cite{chen2018ionet} and RoNIN~\cite{yan2019ronin}) have shown that integrating average velocity estimates from a trained neural network results in highly accurate 2D trajectory reconstruction using only IMU data from pedestrian hand-carried devices. These results showed the ability of networks to learn a translation motion model from pedestrian data. In this work, we draw inspiration from these new findings to build an IMU-only dead reckoning system trained with data collected from a head-mounted device, similar to a VR headset. This paper has two major contributions:

\begin{enumerate}
\item
We propose a network design to regress both the 3D displacement and the corresponding covariance, and show the network's capability of providing both accurate and statistically consistent outputs trained on our pedestrian dataset.
\item
We propose a complete state estimation system combining the neural network with an EKF in a tightly-coupled formulation that jointly estimates position, orientation, velocity, and IMU biases with only pedestrian IMU data.
\end{enumerate}

Our approach presents significant advantages over other IMU-only state estimation approaches. Comparing to traditional PDR methods, it avoids the limitations and complexities of step-counting/stride or gait detection by learning a displacement motion model valid for any data segments, simplifying the process and improving accuracy and robustness. Comparing to common deep learning approaches like RoNIN, it elevates the problem onto 3D domain and does not require an external ZUPT-based orientation estimator. This tight fusion approach reduces average yaw and position drift by 27\% and 33\% respectively on our test dataset comparing to the best performing RoNIN velocity concatenation baseline approach.
Fig.~\ref{fig:overview} shows an example of the estimated trajectory. We name our method \textsc{Tight Learned Inertial Odometry} (\FILTERMETHOD).

\section{Related Works}

\textbf{VIO.}
Visual-Inertial Odometry has been well studied in the literature~\cite{chen2018review}. 
Without power and privacy issues and when static visual features are clearly trackable, VIO is the standard method for state estimation.

\textbf{PDR.}
Solutions to Pedestrian Dead Reckoning problems can be divided into two categories: Bayesian filtering~\cite{jimenez2010indoor} and step counting~\cite{brajdic2013walk}.
With IMU as the only source of information for both propagation and update steps of an EKF, the measurements need to be carefully constructed. Different types of pseudo measurement approaches include Zero-velocity UPdaTe (ZUPT)~\cite{foxlin2005pedestrian} and Zero Angular Rate Update (ZARU)~\cite{rajagopal2008personal} which detects when the system is static, and heuristic heading reduction (HDR)~\cite{borenstein2009heuristic} which identifies walking in a straight line to calibrate gyroscope bias. Foot-mounted IMU sensor enables reliable detection of contact from which accurate velocity constraints can be derived~\cite{ju2016foot}, however such information is not present for hand carried devices, and various signal processing and machine learning algorithms such as peak detection in time/frequency domain and activity classification have been explored~\cite{brajdic2013walk}. These algorithms are complex and involve lots of tuning to accurately estimate a trajectory. It is also an active research direction as deep learning approaches solving for subproblems such as gait classification~\cite{dehzangi2017imu} and stride detection~\cite{beaufils2019robust} are investigated.

\textbf{Deep Learning.} 
The emergence of deep learning provides new possibilities to extract information from IMU data. 
Estimating average velocity from IMU data segments has seen great success in position estimation.
IONet\cite{chen2018ionet} first proposed an LSTM structure to output relative displacement in 2D polar coordinates and concatenate to obtain position. RIDI~\cite{yan2018ridi} and RoNIN~\cite{yan2019ronin} both assume orientation is known from an Android device to rotate IMU data into a gravity-aligned frame. While RIDI regresses velocity to optimize bias but still uses double integration from the corrected IMU data for position, RoNIN directly integrates the regressed velocity and showed better results. This shows that a statistical IMU motion model can outperform the traditional kinematic IMU model in scenarios that can be captured with training data.

In addition to using networks alone for pose estimates, Backprop KF~\cite{haarnoja2016backprop} proposes an end-to-end differentiable Kalman filter framework. Because the loss function is on the accuracy of the filter outputs, the noise parameters are trained to produce the best state estimate, and do not necessarily best capture the measurement error model. AI-IMU~\cite{brossard2019ai} uses this approach to estimate IMU noise parameters and measurement uncertainties for car applications. In this work, we also combine deep learning with a Kalman filter. However, instead of training an end-to-end system, we leverage a state-of-the-art visual-inertial fusion algorithm as ground-truth for supervised learning of the measurement function itself. We use a likelihood loss function to jointly train for 3D displacement and uncertainty, which are directly used as input to the EKF in the measurement update.


 \label{sec:rel}

\section{System Design} \label{sec:sys}

\begin{figure}
	\centering
	\includegraphics[width=0.9\linewidth]{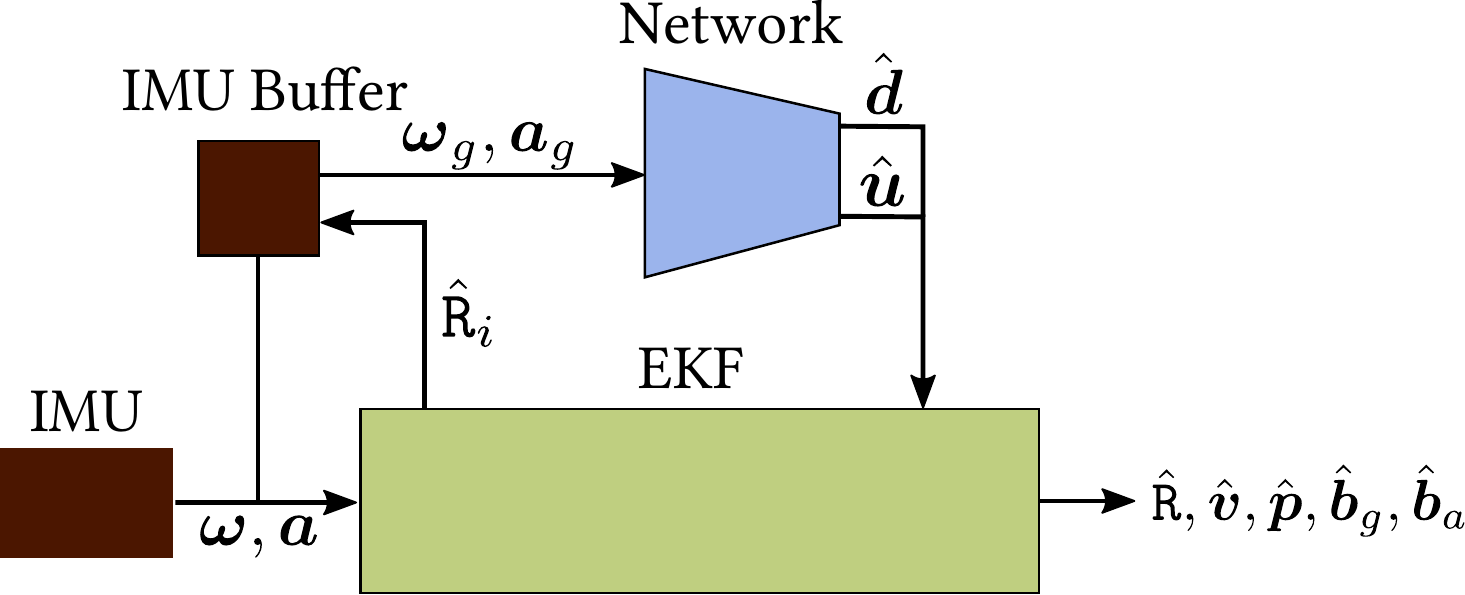}
	\caption{System block diagram. The IMU buffer provides segments of gravity-aligned IMU measurements to the network, using rotation from the filter state. The network outputs displacement $\hat{\dispmeas}$ and uncertainty $\uncertvec$ used as measurement update to the filter. The filter estimates rotation, velocity, position and IMU biases at IMU rate.}
	\label{fig:system}
\end{figure}

Our estimation system takes IMU data as input and has two major components: the network and the filter. Figure~\ref{fig:system} shows a high-level diagram of the system.

The first component is a convolutional neural network trained to regress 3D relative displacement and uncertainty between two time instants given the segment of IMU data in between. 
The network is required to infer positional displacement over a short timespan from acceleration and angular velocity measurements alone, without access to the initial velocity. 
This is intentional: It forces the network to learn only a prior expressing "typical human motion" and leaves model-based state propagation, i.e. acceleration integration to propagate velocity, and respective uncertainty propagation, to the second system component, the EKF. 
In order for this to work well, we found that it is important to include the inferred prior's uncertainty to the network output, allowing the network to encode how much motion model prior it obtained from the input measurements.

The EKF estimates the current state: 3D position, velocity, orientation and IMU biases, and a sparse set of past poses. 
The EKF propagates with raw IMU samples and uses network outputs for measurement updates. We define the measurement in a local \textit{gravity-aligned frame} to decouple global yaw information from the relative state measurement (see Sec.~\ref{sec:kf-meas}). The propagation from raw IMU data provides a model-based kinematic motion model, and the neural network provides a statistical motion model. The filter tightly couples these two sources of information.

At runtime, the raw IMU samples are interpolated to the network input frequency and rotated to a local gravity-aligned frame using rotations estimated from the filter state and gyroscope data. A gravity-aligned frame always has gravity pointing downward.
Placing data in this frame implicitly gives the gravity direction information to the network.

Note that in the proposed approach, IMU data is being used twice, both as direct input for state propagation and indirectly through the network as the measurement. This violates the independence assumptions the EKF is based on:
the errors in the IMU network input (initial orientation, sensor bias, and sensor noise) would propagate to its output which is used to correct the propagation results polluted with the same noise, which could lead to estimation inconsistency.
We address this issue with training techniques (see Sec.~\ref{sec:net}) to reduce this error propagation by adding random perturbations to IMU bias and gravity direction during training. We show in Section~\ref{sec:abla} that these techniques successfully improved the robustness of the network to sensor bias and rotation inaccuracies.

\section{Neural Statistical Motion Model} \label{sec:net}
\subsection{Architecture and Loss Function Design}

Our network uses a 1D version of ResNet18 architecture proposed in \cite{he2016deep}. The input dimension to the network is $N\times 6$, consisting of $N$ IMU samples in the gravity-aligned frame. The output of the network contains two 3D vectors: the displacement estimates $\hat{\dispmeas}$ and their uncertainties $\uncertvec$ which parametrize the diagonal entries of the covariance. The two vectors have independent fully-connected blocks extending the main ResNet architecture. 

We make use of two different loss functions during training: the Mean Square Error 
(MSE) and the Gaussian Maximum Likelihood loss. 
The MSE loss on the trained dataset is defined as:
\begin{align}
\mathcal{L}_{\text{MSE}}(\dispmeas, \hat{\dispmeas}) = \frac{1}{n}\sum_{i=1}^n{\rVert \dispmeas_i-\hat{\dispmeas}_i\rVert^2}
\end{align}
where $\hat{\dispmeas} = \{\hat{\dispmeas}_i\}_{i\le n}$ are the 3D displacement output of the network and $\dispmeas = 
\{\dispmeas_i\}_{i\le n}$ are the 
ground truth displacement. $n$ is the number of data in the training set.

We define the Maximum Likelihood loss as the negative log-likelihood of the displacement according to the regressed Gaussian distribution:
\begin{align}
\label{eq:mle-loss}
\mathcal{L}_{\text{ML}}&(\dispmeas,\hat{\Cov},\hat{\dispmeas}) = \frac{1}{n}\sum_{i=1}^n-\log\left(
\tfrac{1}{\sqrt{8\pi^3\det(\hat{\Cov}_i)}}
e^{-\frac{1}{2}\left\rVert{\dispmeas_i-\hat{\dispmeas}_i}\right\rVert^2_{\hat{\Cov}_i}}
\right) \nonumber \\
&= \frac{1}{n}\sum_{i=1}^n
\left(\tfrac{1}{2}\log\det(\hat{\Cov}_i)+\tfrac{1}{2}\rVert{\dispmeas_i-\hat{\dispmeas}_i}\rVert_{{\hat{\Cov}_i}}^2\right) + 
\text{Cst}
\end{align}
where $\hat{\Cov} = \{\hat{\Cov}_i\}_{i\le n}$ are the $3\times 3$ covariance matrices for $i^{th}$ data as a function of the network uncertainty output vector $\uncertvec_i$.
$\hat{\Cov}_i$ has 6 degrees of freedom, and there are various covariance parametrizations for neural network uncertainty estimation~\cite{russell2019multivariate}. In this paper, we simply assume a diagonal covariance output, parametrized by 3 coefficients written as: 
\begin{align}
\hat{\Cov}_i(\uncertvec_i) = \text{diag}(e^{2\uncertvec_{xi}}, e^{2\uncertvec_{yi}}, e^{2\uncertvec_{zi}})
\end{align}
The diagonal assumption decouples each axis, while regressing the logarithm of the standard deviations removes the singularity around zero in the loss function, adding numerical stabilization and helping the convergence in the optimization process.
This choice constrains the principal axis of the uncertainty ellipses to be along the gravity-aligned frame axis.

\subsection{Data Collection and Implementation Details}
\label{sec:net-impl}

We use a dataset collected by a custom rig where an IMU (Bosch BMI055) is mounted on a headset rigidly attached to the cameras. The full dataset contains more than 400 sequences totaling 60 hours of pedestrian data that pictures a variety of activities including walking, standing still, organizing the kitchen, playing pool, going up and down the stairs etc. It was captured with multiple different physical devices by more than 5 people to depict a wide range of individual motion patterns and IMU systematic errors. 
A state-of-the-art visual-inertial filter based on \cite{mourikis2007multi} provides position estimates at 1000 Hz on the entire dataset. We use these results both as supervision data in the training set and as ground truth in the test set. The dataset is split into $80\%$ training, $10\%$ validation and $10\%$ test subsets randomly. 

For network training, we use an overlapping sliding window on each sequence to collect input samples. Each window contains $N$ IMU samples of total size $N\times 6$. In our final system we choose $N=200$ for \SI{200}{Hz} IMU data. We want the network to capture a motion model with respect to the gravity-aligned IMU frame, therefore the IMU samples in each window are rotated from the IMU frame to a gravity-aligned frame built from the orientation at the beginning of the window. We use visual-inertial ground-truth rotation for that purpose.
The supervision data for the network output is computed as the difference of the ground-truth position between two instants expressed in the same headset-centered, gravity-aligned frame.

During training, because we assume the headset can be worn at an arbitrary heading angle with respect to the walking direction, we augment the input data for the network to be yaw invariant by giving a random horizontal rotation to each sample following RoNIN~\cite{yan2019ronin}. 
In our final estimator, the network is fed with potentially inaccurate input from the filter, especially at the initialization stage. We simulate this at training time by random perturbations on the sensor bias and the gravity direction to reduce network sensitivity to these input errors. To simulate bias, we generate additive bias vectors with each component independently sampled from uniform distribution in $[-0.2,0.2]\, m/s^{2}$ or $[-0.05,0.05]\,rad/s$ for each input sample. Gravity direction is perturbed by rotating those samples along a random horizontal rotation axis with magnitude sampled from $[0,5^{\rm o}]$.

Optimization is done through the Adam optimizer. We used an initial learning rate of $0.0001$, zero weight decay, and dropouts with a probability of $0.5$ for the fully connected layers. We observe that training for $\mathcal{L}_{\text{ML}}$ directly does not converge. Therefore we first train with $\mathcal{L}_{\text{MSE}}$ for 10 epochs until the network stabilizes, then switch to $\mathcal{L}_{\text{ML}}$ until the network fully converges. It takes around another 10 epochs to converge and a total of 4 hours of training time is needed on an NVIDIA DGX computer.

\section{Stochastic Cloning Extended Kalman Filter} \label{sec:kf}

The EKF in our system tightly integrates the displacements predicted by the network with a statistical IMU model as used in other inertial navigation systems. As the displacement estimates from the network express constraints on pairs of past states, we adopt a stochastic cloning framework~\cite{roumeliotis2002stochastic}. Similar to~\cite{mourikis2007multi}, we maintain a sliding window of $m$ poses in the filter state. In contrast to~\cite{mourikis2007multi}, however, we only apply constraints between {\em pairs} of poses, and these constraints are derived from the network described in the preceding section, rather than camera data.

\subsection{State Definition} \label{sec:kf-state}

At each instant, the full state of the EKF is defined as:
\begin{align*}
\State = (\clst_1, \dots, \clst_m, \cust)
\end{align*}
where $\clst_i$, $i=1,\dots, m$ are past (cloned) states, and $\cust$ is the current state. More specifically,
\begin{align*}
\clst_i = (\rot_i,  \tran_i), \quad \cust = (\rot, \speed, \tran, \bias_g, \bias_a)
\end{align*}
We express $\rot$ as the rotation matrix that transforms a point from the IMU frame to the world frame, and $\speed$ and $\tran$ are respectively the velocity and position expressed in the world frame. In the following, we will drop the explicit superscript for conciseness.
$\bias_g, \bias_a$ are the IMU gyroscope and accelerometer biases.

As commonly done in such setups, we apply the error-based filtering methodology to linearize locally on the manifold of the minimal parametrization of the rotation. More specifically, the filter covariance is defined as the covariance of the following error-state:
\begin{align*}
\tilde{\clst}_i = (\errr_i, \errp_i), \quad \tilde{\cust} = (\errr, \errv, \errp, \errb_g, \errb_a)
\end{align*}
where \textit{tilde} indicates errors for every state as the difference between the estimate and the real value, except for rotation error which is defined as $\errr = \log_{SO3}(\R \hat{\R}^{-1}) \in \sof(3)$ where $\log_{SO3}$ denotes the logarithm map of rotation. 
The full error-state is of dimension $6m+15$, where $m$ is the total number of past states, and $\tilde{\cust}$ has dimension of $15$.

\subsection{IMU model} \label{sec:kf-imu}
We assume that the IMU sensor provides direct measurements of the non-gravitational acceleration $\acc$ and angular velocity $\rotvel$, expressed in the IMU frame. We assume as it is common for this class of sensor that the measurements are polluted with noise $\nb_{g;a}$ and bias $\bias_{g:a}$.
\begin{align*}
\rotvel = \rotvel_{\text{true}} + \bias_g + \nb_g,%
\quad \acc = \acc_{\text{true}} + \bias_a + \nb_a.
\end{align*}
$\nb_g$ and $\nb_a$ are random noise variables following a zero-centered Gaussian distribution. Evolution of biases is modeled as a random walk process with discrete parameters $\brw_{gd}$ and $\brw_{ad}$ over the IMU sampling period $\Delta t$.%

\subsection{State Propagation and Augmentation} \label{sec:kf-prop}
The filter propagates the current state $\cust$ with IMU data using a kinematic motion model. If the current timestamp is associated to a measurement update, stochastic cloning is performed together with propagation in one step. During cloning, a new state $\clst$ is appended to the past state.

\subsubsection{Propagation Model}
We use the strapdown inertial kinematic equation assuming a uniform gravity field $\grav$ and ignoring Coriolis forces and the earth's curvature:
\begin{gather}
\rotest_{k+1} = \rotest_k \,\exp_{SO3}((\rotvel_k - \hat{\bias}_{gk}) \Delta t)\\
\speedest_{k+1} = \speedest_k +\grav\Delta t + \rotest_k(\acc_k-\hat{\bias}_{ak})\Delta t \\
\tranest_{k+1} = \tranest_k + \speedest_k \Delta t + {\tfrac{1}{2}} \Delta t^2 ( \grav + \rotest_k(\acc_k-\hat{\bias}_{ak}))\\
\hat{\bias}_{g(k+1)} = \hat{\bias}_{gk} + \brw_{gdk}\\
\hat{\bias}_{a(k+1)} = \hat{\bias}_{ak} + \brw_{adk}\text{.}
\end{gather}
Here, $\exp_{SO3}$ denotes the SO(3) exponential map, the inverse function of $\log_{SO3}$.
The discrete-time noise $\brw_{gdk}$, $\brw_{adk}$ have been defined above and follow a Gaussian distribution.

The linearized error propagation can be written as:
\begin{equation}
\tilde{\cust}_{k+1} = \Js_{k(15,15)}^s \tilde{\cust}_k + \Ju_{k(15,12)}^s \nb_k \label{eq:kf-prop}
\end{equation}
where $\nb_{k} = [\nb_{\omega k}, \nb_{ak}, \brw_{gdk}, \brw_{adk}]^T$ is a vector containing all random input. The superscript $s$ indicates that these matrices correspond to the current state $\cust$, and the subscript brackets indicate matrix dimensions.
The corresponding propagation of the state covariance $\Pb$ is:
\begin{gather}
\Pb_{k+1} = \Js_k \Pb_k \Js_k^T + \Ju_k \Wb \Ju_k^T,\\
\Js_k = \begin{bmatrix}
\Iv_{6m}&\bm{0}\\
\bm{0}&\Js_k^s \end{bmatrix}, \Ju_k = \begin{bmatrix}
\bm{0}\\\Ju_k^s
\end{bmatrix}
\end{gather}
with $\Wb_{(12,12)}$ the covariance matrices of sensor noise and bias random walk noise. The state covariance $\Pb$ is of the dimension $(6m+15)$ of the full state $\State$. In the implementation, multiple propagation steps can be combined together to reduce the computational cost~\cite{roumeliotis2002stochastic}.

\subsubsection{State Augmentation}

In our system, state augmentations are performed at the measurement update frequency. During an augmentation step, the state dimension is incremented through propagation with cloning:
\begin{gather}
\Pb_{k+1} = \bar{\Js}_k \Pb_k \bar{\Js}_k^T + \bar{\Ju}_k \Wb \bar{\Ju}_k^T,\\
\bar{\Js}_k = \begin{bmatrix}
\Iv_{6m}&\bm{0}\\
\bm{0}& {\Js}^{\clstexp}_k\\
\bm{0}&{\Js}_k^s \end{bmatrix}, \bar{\Ju}_k = \begin{bmatrix}
\bm{0}\\\Ju^{\clstexp}_k\\\Ju_k^s
\end{bmatrix}
\end{gather}
$\bar{\Js}_k$ is now a copy operation plus the augmented and current state propagation operations. $\Js^{\clstexp}_k$ and $\Ju^{\clstexp}_k$ are partial propagation matrices for rotation and position only. After the increment, the dimension of the state vector increases by 6. Old past states will be pruned in the marginalization step, see Sec.\ref{sec:marginit}. 

\subsection{Measurement Update} \label{sec:kf-meas}

It would be natural to define the measurement function  as the displacement $\dispmeas$ expressed in the world frame, however such measurement function would imply heading observability at the filter level. Heading is theoretically unobservable as the IMU propagation model and the learned prior are invariant to the change of yaw angle. In order to prevent injecting spurious information into the filter, we carefully define the measurement function $h$ as the 3D displacement in a \textit{local gravity-aligned} frame. This frame is anchored to a clone state $i$. Vectors in this frame can be obtained by rotating the corresponding world frame vectors by the yaw rotation matrix ${\R}_{\gamma}$ of the state rotation matrix ${\R}_i$. We decompose ${\R}_i$ using extrinsic "XYZ" Euler angle convention: ${\R}_i = {\R}_{\gamma} {\R}_{\beta} {\R}_{\alpha}$, where $\alpha$, $\beta$, $\gamma$ correspond to roll, pitch and yaw respectively. $h$ then writes:
\begin{equation}
h(\State) = {\R}_{\gamma}^T({\tran}_j-{\tran}_i) = \dispest_{ij} + \brw_{{\dispmeas}_{ij}}.
\label{eq:kf-meas}
\end{equation} 
$\dispest_{ij}$ is the network output displacement between state $i$ and $j$, using IMU samples rotated to the local gravity-aligned frame anchored at pose $i$ as input. $\brw_{{d}_{ij}}$ is a random variable that, we assume, follows the normal distribution $\mathcal{N}(0, \hat{\Cov}_{ij})$ given by the network.

The linearization of the measurement function $h$ with respect to the error state yields the linearized measurement matrix $\Hv_{(3,\,6m+15)}$. It has only non zero values in the $3\times 3$ blocks corresponding to $\errr_i$, $\errp_i$ and $\errp_j$. 
\begingroup
\setlength\arraycolsep{3pt}
\begin{align}
&\Hb_{\errr_i} = \frac{\partial h(\State)}{\partial \errr_i} = \hat{\R}_{\gamma}^T \floor{\tranest_j-\tranest_i}_{\times} \Hv_z\\ 
&\Hb_{\errp_i} = \frac{\partial h(\State)}{\partial \errp_i} = -\hat{\R}_{\gamma}^T\\
&\Hb_{\errp_j} = \frac{\partial h(\State)}{\partial \errp_j} = \hat{\R}_{\gamma}^T\\
&\text{ where } \Hb_z = \begin{bmatrix}
0 & 0 & 0\\
0 & 0 & 0\\
\cos{\gamma}\tan{\beta} & \sin{\gamma}\tan{\beta} & 1
\end{bmatrix}
\end{align}
\endgroup
In the above expression, $\floor{\xb}_\times$ is the skew-symmetric matrix built from vector $\xb$.
The singularity $\cos{\beta}=0$ occurs when the person wearing the headset is looking straight down or straight up: we simply discard the update for these cases. Note that these situations are unlikely and never occur in our dataset.

Finally, $\Hb$ and $\hat{\Cov}_{ij}$ are used to compute the Kalman gain to update the state $\State$ and the covariance $\Pb$ as follows:
\begin{align}
\Kb &= \Pb\Hb^T(\Hb\Pb\Hb^T + \hat{\Cov}_{ij})^{-1}\\
\State &\longleftarrow \State\oplus \left(\Kb(h(\State)-\dispest_{ij})\right)
\\
\Pb &\longleftarrow (\Ib-\Kb\Hb)\Pb(\Ib-\Kb\Hb)^T + \Kb\hat{\Cov}_{ij}\Kb^T
\end{align}
Operator $\oplus$ denotes the regular addition operation except for rotation where the update operation writes $\R \leftarrow \text{Exp}(\errr)\R$.

We use a $\chi_2$ test to protect the filter estimate against occasional wrong network output: we discard the update when the normalized innovation error $\|\dispmeas_{ij}-\dispest_{ij}\|_{\Hb\Pb\Hb^T + \hat{\Cov}_{ij}}$ is greater than a threshold. We choose here the threshold value of 11.345 which corresponds to the $99^{th}$ percentile of the $\chi_2$ distribution with 3 degrees of freedom.

In practice, when using the measurement covariance $\hat{\Cov}_{ij}$ from the network, we scale the covariance by a factor of 10 to compensate for the temporal correlation of measurements as noted in Sec.~\ref{sec:PastDataWindowSize}.

\subsection{State size, Marginalization and Initialization} \label{sec:marginit}
As soon as an update is processed, all states prior to this update window are marginalized. This is performed by removing the corresponding $\clst$ from the state and the corresponding covariance entries.
The number of states kept in the filter depends on two parameters: (i) the displacement duration window for which the network is trained and (ii) the update frequency. For instance, for an update frequency of \SI{20}{Hz} and a window of \SI{1}{s}, there will be at most 21 past states in the filter. The two parameters can be set independently: if the period of update is smaller than the displacement window, the consecutive measurements will be generated from the overlapping windows of IMU data, while if the update period is too long, some IMU measurements would not be given to the network. We observe that our system has better performance with a higher update frequency (see Sec.~\ref{sec:ekf-tuning}), and we use \SI{20}{Hz} in our final system.

Our EKF needs a good initialization to converge. In this work we assumed the initial state is given by an external procedure. In our experiments we initialize the speed, roll, and pitch with the ground-truth estimate assigning a large covariance, while the yaw and the position are initialized with a strong prior in order to fix the gauge. Biases are initialized as with the values of an initial factory calibration, and we refine the biases online to account for its evolution due to turn-on change, temperature effect, or aging.

In practice, we use the following values to initialize the state covariance:
$\sigma_{\speed}=0.1$ m/s, $\sigma_{\bias_a}=0.2$ m/s$^2$, $\sigma_{\bias_g}=1.10^{-4}$ rad/s, $\sigma_{\theta}=\text{diag}(10,10,0.1)$ deg.
We also compensate all the input IMU samples for non-orthogonality scale factor and g-sensitivity from the factory calibration.

\section{Experimental Results} \label{sec:res}
We evaluate our system on the test split of our dataset. Each of these 37 trajectories contains between 3 to 7 minutes of human activities. We compare the performance of two pose estimators against a state-of-the-art VIO implementation based on \cite{mourikis2007multi} and considered as ground-truth (GT):
\begin{itemize}
    \item \FILTERMETHOD: the tightly-coupled EKF with displacement update from the trained network presented in this work.
    \item \RONINMETHOD: an estimator where the displacements from the same trained network are concatenated in the direction given by an engineered AHRS attitude filter, resembling what smartphones have. We borrow the name from \cite{yan2019ronin}, but we extended the method to 3D and we trained the network on our dataset to get a fair comparison. Note that our dataset contains trajectories involving climbing staircases, sitting motions, and walking outdoors on uneven terrain, where the 2D PDR method of \cite{yan2019ronin} is not directly applicable.
\end{itemize}

\begin{figure}
	\centering
	\includegraphics[trim=75 5 75 5, clip=true, width=\linewidth]{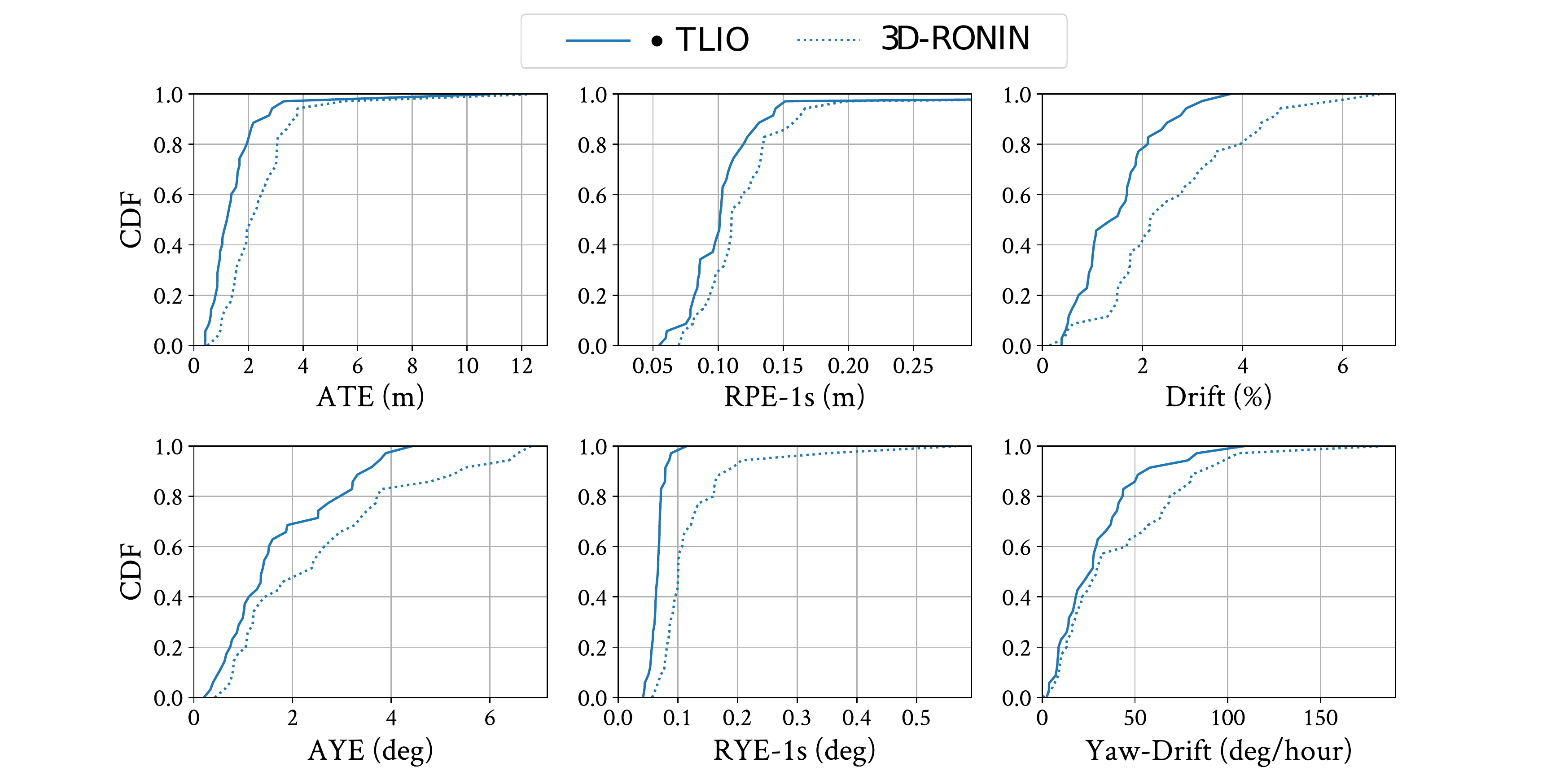}
	\caption{Comparing \FILTERMETHOD to \RONINMETHOD on the error metrics with respect to the ground-truth. Each plot shows the cumulative density function of the chosen metric on the entire test set. The steeper the plots are the better the performance. \FILTERMETHOD shows significantly lower value than \RONINMETHOD for all presented metrics.
	}
	\label{fig:statsSystem}
\end{figure}

\begin{figure}
	\centering
	\includegraphics[width=1\linewidth]{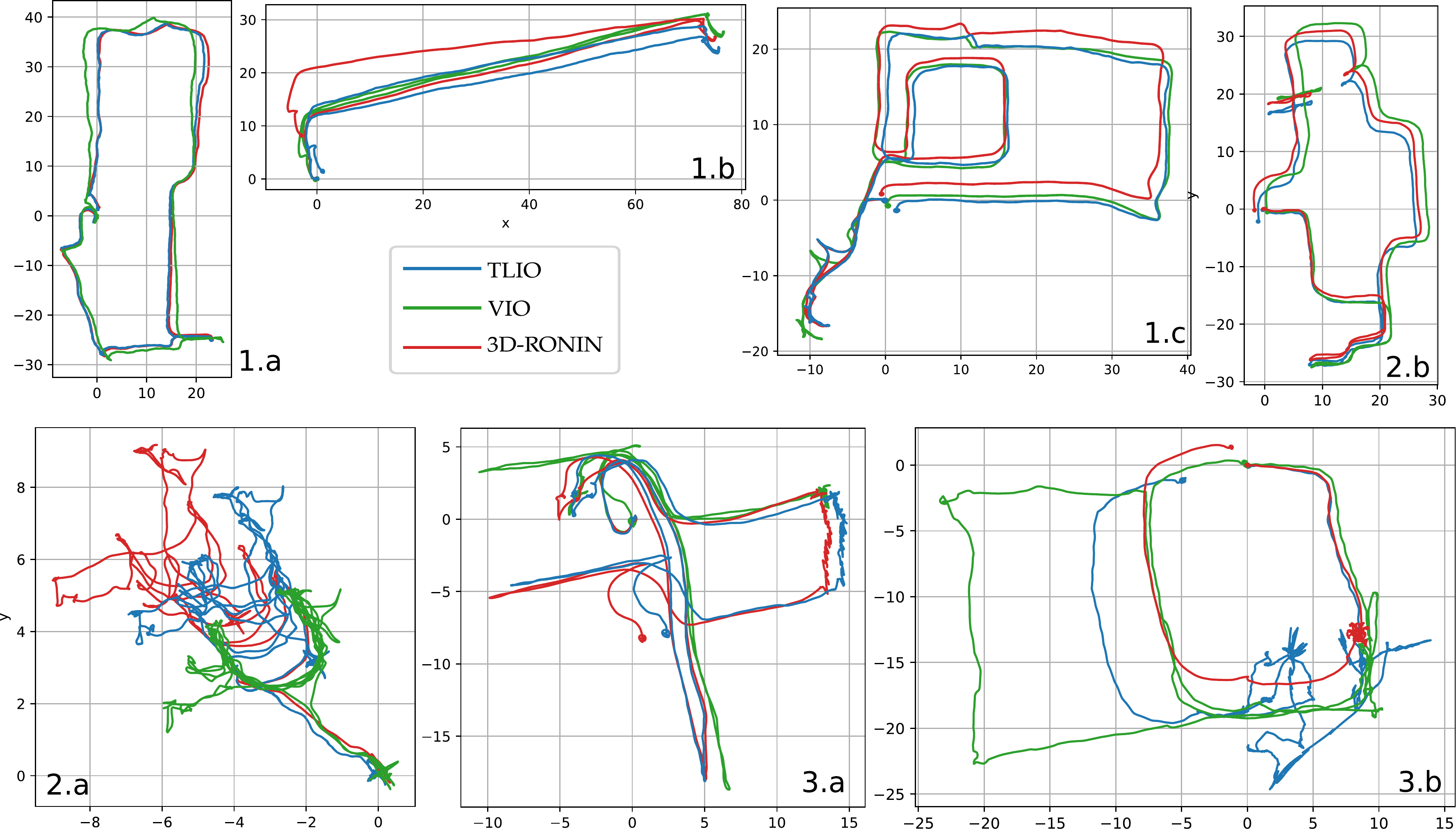}
	\caption{Selection of trajectories. 1.x and 2.x are the good and hard cases and 3.x show examples of failures. 2.a. shows a case with very wrong network outputs. 2.b shows a case of bad initialization in the filter leading to yaw drift. In 3.a and 3.b, unusual motions not present in the training set are performed: side-stepping repeatedly for 3 minutes and rolling on a chair.
	}
	\label{fig:exampleTraj}
\end{figure}

\subsection{Metrics definitions}
In order to assess our system performance, for each dataset of length $n$, we define the following metrics:
\begin{itemize}
\item{\textbf{ATE} (m)}: 
$\sqrt{\frac{1}{n}\sum_i^n \|\tran_i - \tranest_i\|^2}$ 

The Absolute Translation Error indicates the spatial closeness of position estimate to the GT over the whole trajectory, computed as root-mean square error (RMSE) between these sets of values.

\item{\textbf{RTE}-$\Delta t$ (m)}:\\
$\sqrt{\frac{1}{n}\sum_i^n \|\tran_{i+\Delta t} - \tran_{i} - \R_{\gamma}\hat{\R}_{\gamma}^T(\tranest_{i+\Delta t} - \tranest_{i})\|^2}$

The Relative Translation Error indicates the local spatial closeness of position estimate to the GT over a window of duration $\Delta t$. We use \SI{1}{s} in this analysis. We remove the yaw drift at the beginning of the window so that the relative measure is not affected by accumulated errors.

\item{\textbf{DR} (\%)} : 
$(\|\tran_n - \tranest_n\|)/(\text{trajectory-length})$ 

The final translation drift over the distance traveled.
\end{itemize}

We compute similar metrics for yaw angle $\gamma$, measuring the quality of the unobservable orientation around gravity:

\begin{itemize}
\item{\textbf{AYE} (\si{\degree})}: 
$\sqrt{\left(\frac{1}{n}\sum_i \|\gamma_i - \hat{\gamma}_i\|^2\right)}$ is the Absolute Yaw Error.

\item{\textbf{RYE}-$\Delta t$ (\si{\degree})}: 
$\sqrt{\frac{1}{n}\sum_i^n \|\gamma_{i+\Delta t} - \gamma_{i} - (\hat{\gamma}_{i+\Delta t} - \hat{\gamma}_{i})\|^2}$ is the Relative Yaw Error

\item{\textbf{Yaw-DR} (\si{\degree}/hour)}: 
$(\gamma_n - \hat{\gamma}_n)/(\text{sequence-duration})$ 
is the Yaw Drift over time.
\end{itemize}

\subsection{System Performance}


Figure~\ref{fig:statsSystem} shows the distribution of the metrics across the entire test set. It demonstrates that our system consistently performs better than the decoupled orientation and position estimator \RONINMETHOD on all metrics.

On 3D position, \FILTERMETHOD performs better than \RONINMETHOD which integrates average velocities. Integration approach has the advantage of being robust to outliers since the measurements at each instant are decoupled. The result shows not only the benefit of integrating IMU kinematic model, but also the overall robustness of the filter, which comes from the quality of the covariance output from the network and the effectiveness of outlier rejection with $\chi_2$ test.

Our system also has a smaller yaw drift than \RONINMETHOD, indicating that even without any hand engineered heuristics or step detection, using displacement estimates from a trained statistical model outperforms a smartphone AHRS attitude filter. This shows that the EKF can accurately estimate not only position and velocity, but also gyroscope biases.

Figure~\ref{fig:exampleTraj} shows a hand-picked collection of 7 trajectories containing the best, typical and worst cases of \FILTERMETHOD and \RONINMETHOD, while Figure~\ref{fig:overview} shows a 3D visualization of one additional sequence. See the corresponding captions for more details about the trajectories and failure cases.

\section{Components and Variation Studies} \label{sec:abla}

We investigate different variations of the network settings in three aspects: input IMU frequency, time interval $\Delta t_{ij}$ over which displacement is computed, and the total time interval included in the network input. We do not consider future data because we aim for a causal filter. The name convention we use is built from these aspects. For example, \texttt{200hz-05s-3s} is a model that takes \SI{3}{s} of \SI{200}{Hz} input data and regresses for the displacement over the last \SI{0.5}{s}. To easily keep track of the different models, in all following figures, we prefix with ``$\bullet$" the version of our algorithm whose results were presented in Sec.~\ref{sec:res}: \texttt{200hz-1s-1s}.

\subsection{Network Statistical Model}
In this section, we evaluate the deep learning component of our system in isolation.

\subsubsection{Does using more data help?} \label{sec:PastDataWindowSize}

\begin{figure}
	\centering
	\includegraphics[trim=20 0 0 0, clip=true, width=\linewidth]{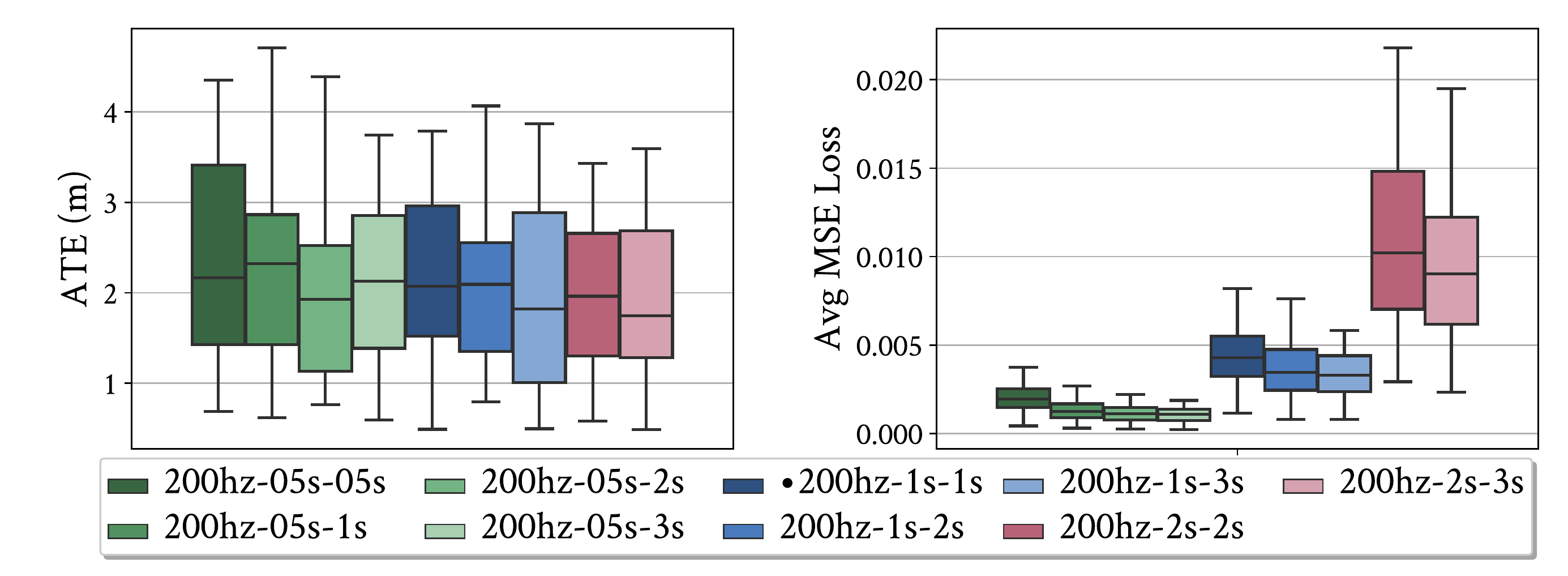}
	\caption{Network variations with different past and total window sizes evaluated on the test set. Upper left: ATE of \RONINMETHOD concatenation result comparing to GT trajectories. Upper right: average MSE loss on the test set. Adding past data to the network input reduces average MSE loss over samples, but it does not imply lower ATE over trajectories.}
	\label{fig:net-vars}
	\begin{minipage}[t]{0.35\linewidth}
	\strut\vspace*{-\baselineskip}\newline\includegraphics[height=3cm]{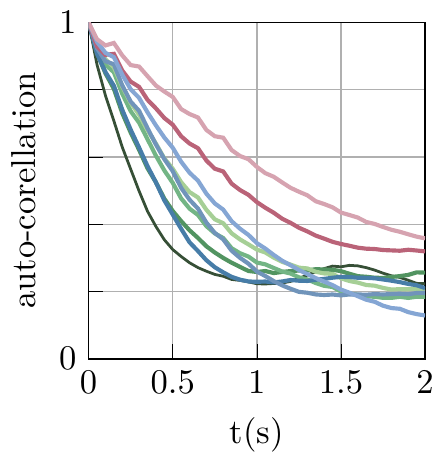}
	\end{minipage}\begin{minipage}[t]{0.65\linewidth}
	\caption*{Left: Auto-correlation function of the norm of the network displacement error over one sequence. The norms are computed at \SI{20}{Hz} using overlapping windows of IMU data. Models with more input data have lower MSE but greater temporal correlation of errors.}
	\end{minipage}
\end{figure}

Figure~\ref{fig:net-vars} shows a comparison of several variations of the window duration, keeping the IMU frequency fixed to \SI{200}{Hz}. It is surprising that lower MSE loss over the same displacement window does not imply lower ATE. This is because MSE does not say anything about the sum of the errors: the network can make more accurate predictions per sample by seeing more data, while at the same time produces more correlated results as the inputs overlap over a longer period of time. This introduces a trade-off when integrating displacement, and we observe similar trajectory ATEs in different network time window variants. For further experiments we will use \texttt{200hz-1s-1s} except noted otherwise.

\subsubsection{Consistency of learned covariance}

\begin{figure}
	\centering
	\includegraphics[width=\linewidth]{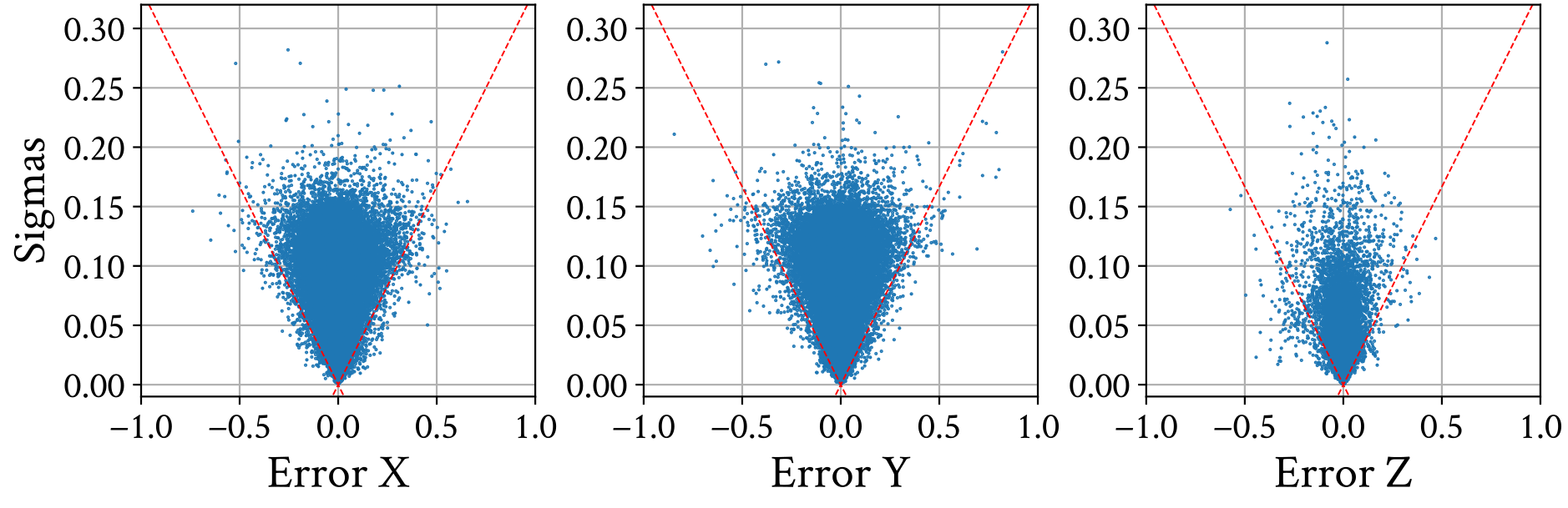}
	\caption{Uncertainty returned by the network (standard-deviation $\sigma$) plotted against errors (m) on three axes $x,y,z$ respectively. Points with larger errors have larger variance outputs, and over $99\%$ of the points fall inside the $3\sigma$ cone region indicated by the dashed red line. We have $0.70\%$ for $x, y$ and $0.47\%$ for $z$ axis of error points outside $3\sigma$ bounds respectively.}
	\label{fig:sigmas}
	\vspace{-1em}
\end{figure}

We collected around $60\,000$ samples at \SI{5}{Hz} on the entire test set for this analysis and the sensitivity analysis section below. 
Fig.~\ref{fig:sigmas} plots the sigma outputs of the network against the errors on the displacement. We observe over $99\%$ of the error points fall inside the $3\sigma$ region, showing the network outputs are consistent statistically. In addition, we compute the Mahalanobis distance of the output displacement errors scaled by the output covariances in 3D.
We found that only $0.30\%$ of the samples were beyond the $99$ percentile critical value of the $\chi_2$ distribution, all of which are from the failure case shown in Fig.~\ref{fig:exampleTraj}.3.b).  From this analysis, we conclude that the uncertainty output of the network grows as expected with its own error albeit being often slightly conservative. We observe an asymmetry between horizontal and vertical errors distribution; along vertical axis errors and sigmas concentrate on lower values. This is not surprising: people usually either walk on a flat ground or on stairs which may make expressing a prior easier on the z axis.

\subsubsection{Sensitivity Analysis}

\begin{figure}
	\centering
	\includegraphics[trim=15 5 10 5, clip=true, width=\linewidth]{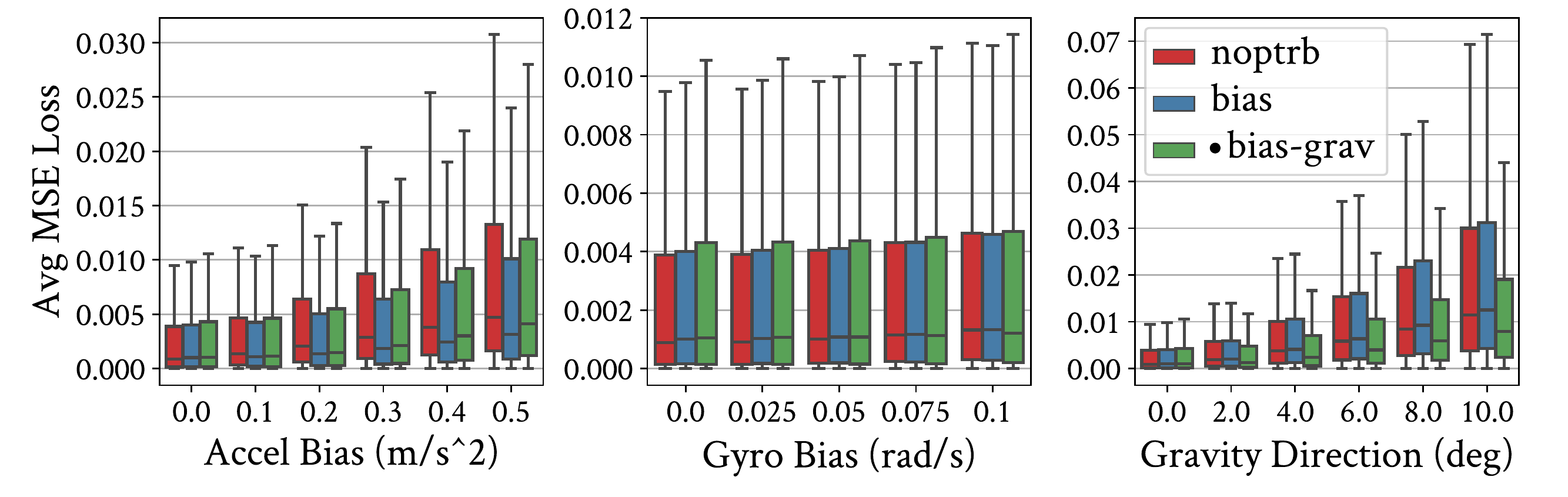}
	\caption{Sensitivity of the network prediction on bias and gravity direction noise. At test time the input IMU samples are perturbed for accelerometer bias, gyroscope bias, and gravity direction as described in Sec.~\ref{sec:net-impl} but with various ranges. The three models under comparison are trained with different data augmentation procedures: no perturbation (noptrb), perturbing only the bias (bias), and perturbing both the bias and the gravity direction (bias-grav). We observe that training with perturbation significantly improves robustness on accelerometer bias and gravity direction.}
	\label{fig:pert}
	\vspace{-1em}
\end{figure}

Figure~\ref{fig:pert} shows the network output robustness to input perturbation on bias and gravity direction. We observe that the models trained with bias perturbation are more robust to IMU bias, in particular accelerometer bias. The model trained with gravity perturbation is significantly more robust to gravity direction errors. These network training techniques effectively help reduce the propagation of errors from input to the output, protecting the independence assumption required in a Kalman Filter framework. Therefore we choose \texttt{200hz-1s-1s} trained with bias and gravity perturbation as our system model.

\subsection{EKF System}

\subsubsection{Ablation Study}
\begin{figure}
	\centering
	\includegraphics[trim=50 20 50 20, clip=true, width=1\linewidth]{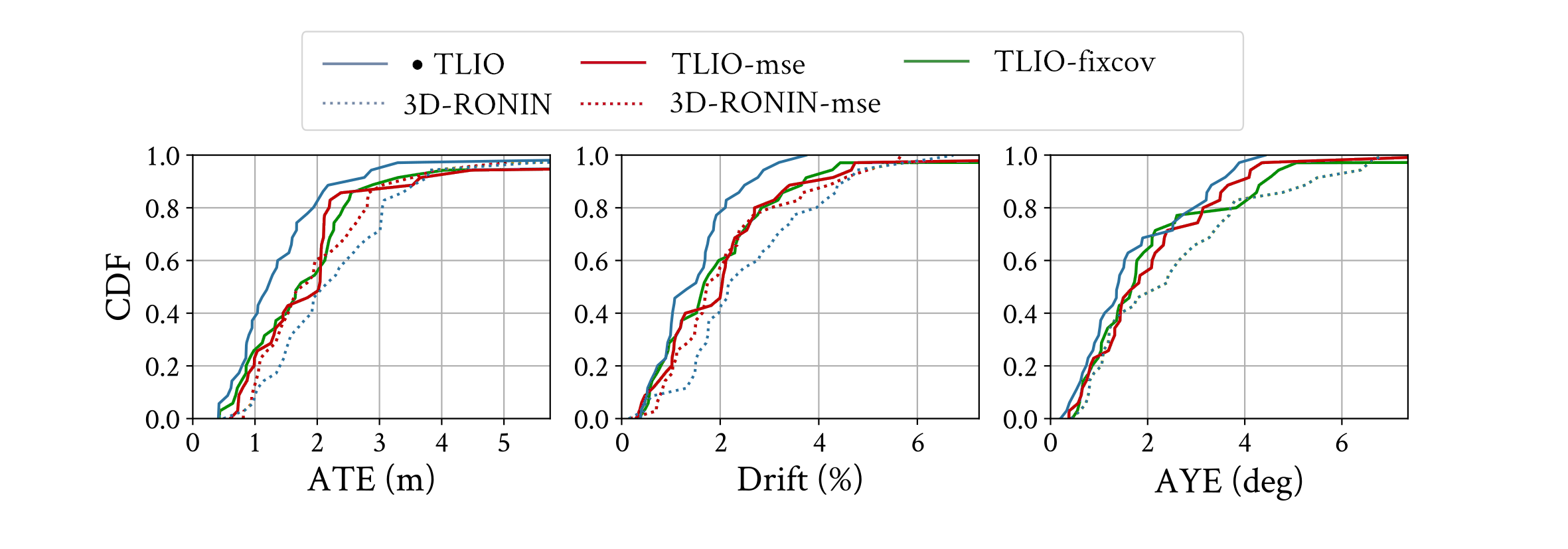}
	\caption{Performance comparison showing the effectiveness of using a learned covariance. \FILTERMETHOD-mse and \RONINMETHOD-mse use a network trained with only MSE loss. \FILTERMETHOD-fixcov and \FILTERMETHOD-mse use a hand-tuned constant covariance for all the measurements. \RONINMETHOD and \RONINMETHOD-mse concatenate displacement estimates directly without considering uncertainty. \FILTERMETHOD achieves the best performance with the covariance regressed from the network. Constant covariance approaches do not reach $100\%$ for ATE and drift in this illustration due to a failure case.
	}
	\label{fig:gravityPerturbResult}
\end{figure}

We compare system variants using a hand-tuned constant covariance and networks trained without likelihood loss to show the benefit of using the regressed uncertainty for \FILTERMETHOD.
To find the best constant covariance parameters, we use the standard deviation of the measurement errors on the test set - $\text{diag}(0.051,0.051,0.013)$m - multiplied by a factor yielding the best median ATE tuned by grid-search.
Fig.~\ref{fig:gravityPerturbResult} shows the Cumulative Density Function of ATE, Drift and AYE over the test set of various system configurations.
We observe that using \RONINMETHOD, training the network with only MSE loss gives a more accurate estimate. However, using a fixed covariance, such network variants in a filter system achieve similar performance. What makes a difference is the regressed covariance from the network, which significantly improves the filter in ATE and Drift.
We also notice a dataset where using a fixed covariance loses track due to initialization failure, showing the adaptive property of the regressed uncertainty for different inputs, improving system robustness. Comparing to \RONINMETHOD-mse, \FILTERMETHOD has an improvement of $27\%$ and $31\%$ on the average yaw and position drift respectively.

\subsubsection{Timing parameters} \label{sec:ekf-tuning}
\begin{figure}
	\centering
	\includegraphics[height=4.5cm]{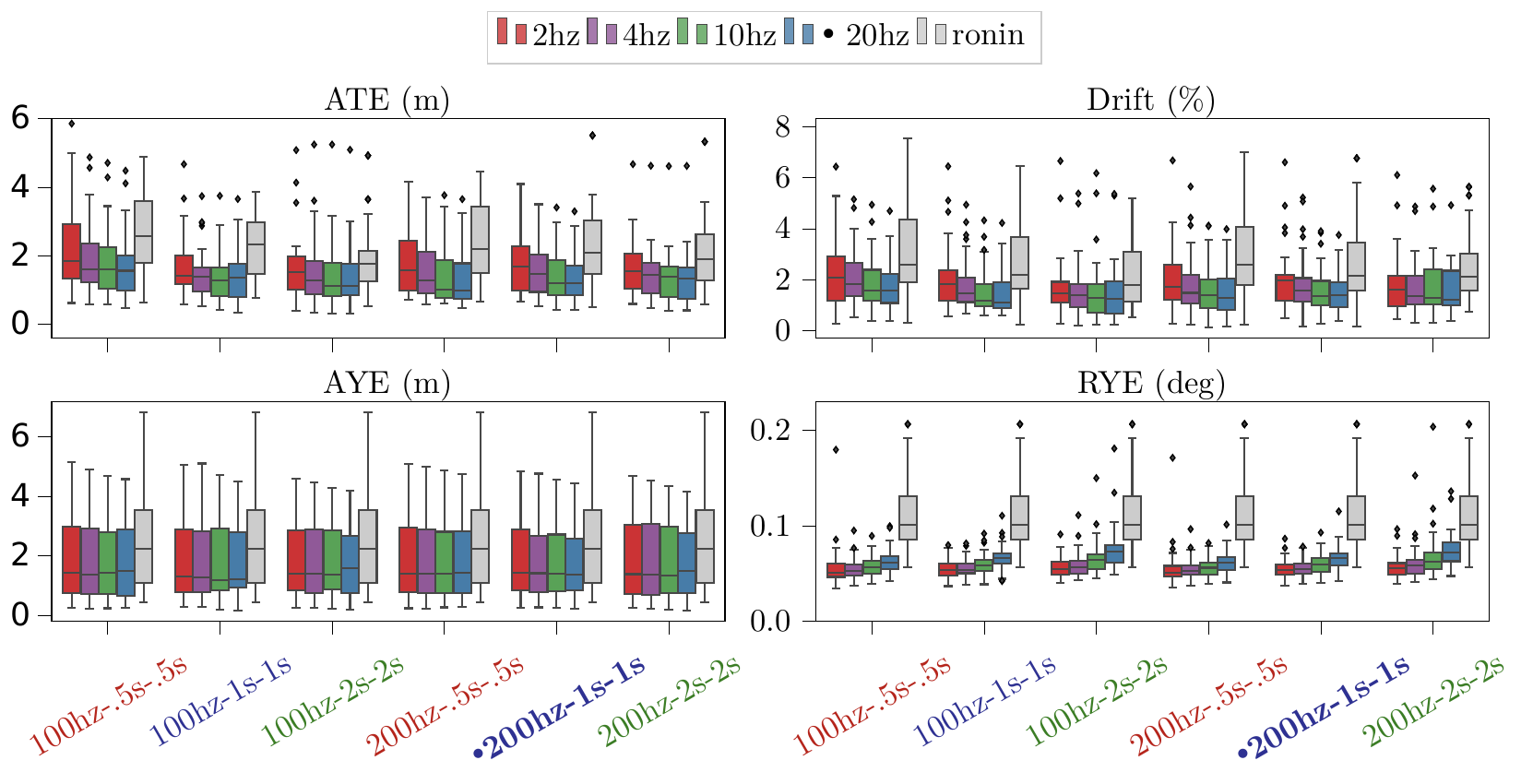}
	\caption{Performance of different system configurations. Each group corresponds to a network model, as indicated by the x axis labels. Within each group, the colored boxplots differ by update frequency, and the grey ones show \RONINMETHOD as baseline. The filtering approach constantly outperforms \RONINMETHOD on all the experimented models. High frequency update yields the best ATE and drift in spite of temporal correlation of the measurements. RYE-1s increases with update frequency, indicating more jitter on the yaw estimates.
	}
	\label{fig:timingParameters}
	\vspace{-1em}
\end{figure}

Fig.~\ref{fig:timingParameters} shows the performance comparison between variations on network IMU frequency, $\Delta t_{ij}$, and filter measurement-update frequency. ATE and drift are reduced when using high frequency measurements. Once again, we observe minor differences between network parameter variations on the monitored metrics. This shows that our entire pipeline is not sensitive to these choices of parameters, and \FILTERMETHOD outperforms \RONINMETHOD in all cases.

\section{Conclusion}

In this paper we propose a tightly-coupled inertial odometry algorithm introducing a learned component in an EKF framework. We present the network regressing 3D displacement and uncertainty, and the EKF fusing the displacement measurement to estimate the full state. We train a network that learns a prior on the displacement distributions given IMU data from statistical motion patterns. We show through experimental results and variation studies that the network outputs are statistically consistent, and the filter outperforms the state-of-the-art trajectory estimator using velocity integration on position estimates, as well as a model-based AHRS attitude filter on orientation. This demonstrates that with a learned prior, an IMU sensor alone can provide enough information to do low drift pose estimation and calibration for pedestrian dead-reckoning. As common with learning approaches, the system is limited by the scope of training data. Unusual motions cause system failure as discussed with examples. Whether similar approach can be generalized to wider use cases such as legged robots is still an unexplored but promising field of research.

\bibliographystyle{IEEEtran}
\bibliography{references}


\end{document}